%% file: main.tex
\documentclass[10pt,twocolumn,letterpaper]{article}

\usepackage[pagenumbers]{wacv} % To force page numbers, e.g. for an arXiv version
\usepackage{svg}
\usepackage{stfloats}
\usepackage{booktabs}
\usepackage{tabularx}
\usepackage{array}
\usepackage{enumitem}
\setlist{nosep}
\svgsetup{
	inkscapeexe={"C:/Program Files/Inkscape/bin/inkscape.com"},
	inkscapeversion=1,
	inkscape=force
}

\usepackage{amsmath}
\usepackage{graphicx}
\usepackage{capt-of}

\definecolor{wacvblue}{rgb}{0.21,0.49,0.74}
\usepackage[breaklinks,colorlinks,allcolors=wacvblue]{hyperref}

\def\wacvPaperID{811} % *** Enter the WACV Paper ID here
\def\confName{WACV}
\def\confYear{2027}

\title{ProGuT: Label-Efficient Panoptic Segmentation for Forest Scenes}

\author{Pankaj Deoli\\
	Robotics Research Lab, RPTU Kaiserslautern-Landau\\
	Gottlieb-Daimler-Str. 48, 67663 Kaiserslautern, Germany\\
	{\tt\small pankaj.deoli@cs.rptu.de}
	\and
	Karsten Berns\\
	Robotics Research Lab, RPTU Kaiserslautern-Landau\\
	Gottlieb-Daimler-Str. 48, 67663 Kaiserslautern, Germany\\
	{\tt\small karsten.berns@cs.rptu.de}
}

\begin{document}
	\twocolumn[{
		\renewcommand\twocolumn[1][]{#1}
		\maketitle
		
		%\vspace{-1.0em}
		\begin{center}
			\includegraphics[width=0.95\textwidth]{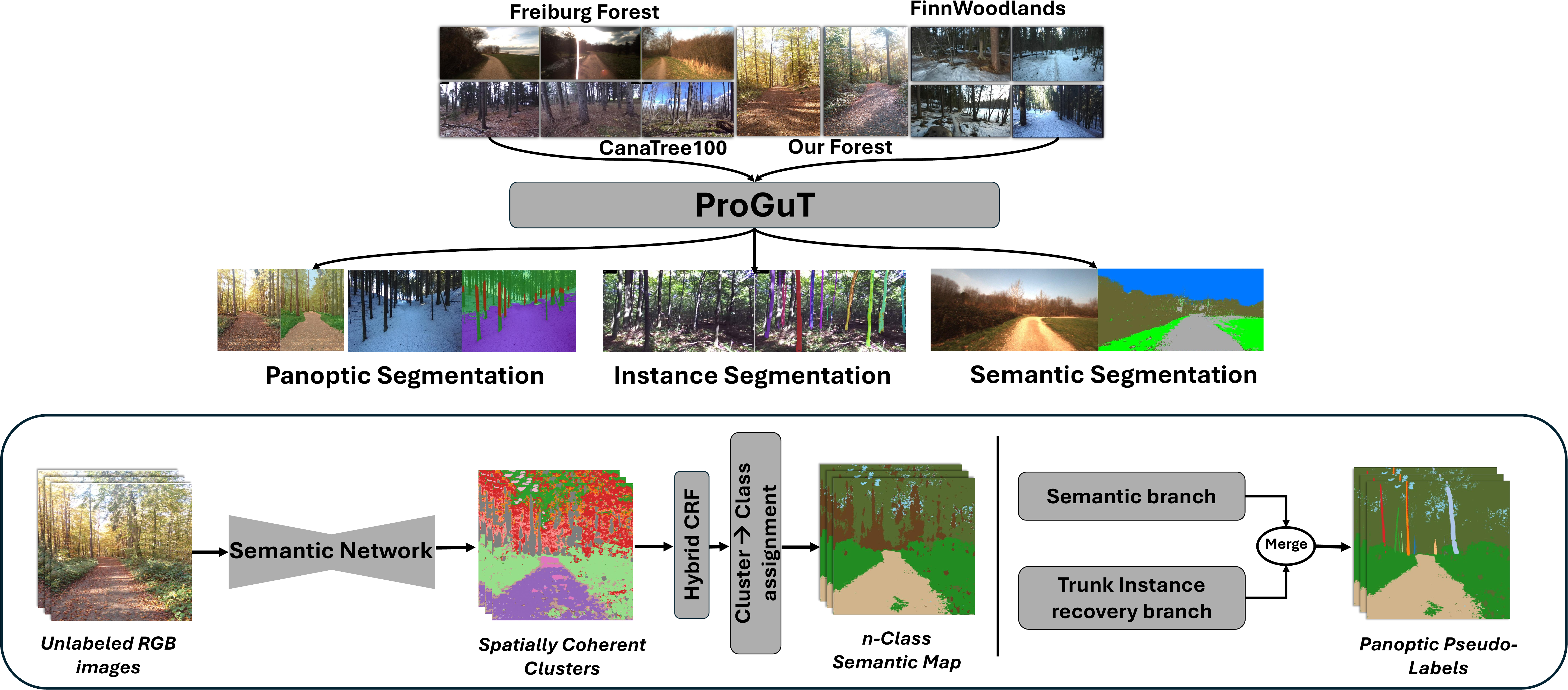}
			\vspace{0.2em}
			\captionof{figure}{Results and overview of \textbf{ProGuT}. \textit{\textbf{Top}:} Given unlabeled forest images, ProGuT produces Semantic, Instance and Panoptic segmentation. \textit{\textbf{Bottom:}} The ProGuT pipeline. A semantic network clusters CLIP patch features  into Spatially Coherent Clusters; followed by a hybrid CRF and cluster-to-class assignment yielding \textit{n}-class semantic map. In parallel, a trunk-instance branch recovers individual trunks via multi-scale geometric prior. The 2 branches are merged to form panoptic pseudo-labels.}
			\label{fig:teaser}
		\end{center}
		\vspace{1.0em}
	}]
	
\input{sec/0_abstract}    
\input{sec/1_intro}
\input{sec/related_works}
\input{sec/methodology}
\input{sec/experimentations}
\input{sec/conclusion}
\clearpage
{
    \small
    \bibliographystyle{ieeenat_fullname}
    \bibliography{main}
}

\end{document}

%% file: sec/0_abstract.tex
\noindent\makebox[\columnwidth][c]{\large\bfseries Abstract}
\par\smallskip

\begingroup
\itshape
\noindent

Panoptic segmentation in forest environments is bottlenecked not by semantic quality but by instance separation; existing unsupervised panoptic approaches produce usable stuff maps but near-zero thing quality. Depth or flow-based instance discovery methods needs sensors that are not always available. We present \textbf{ProGuT (Prototype Guided Training)}, which produces panoptic pseudo-labels without per-image training masks, needing only unlabeled images and one-time cluster-to-class mapping. ProGuT clusters CLIP patch features, then recovers trunk instances through multiscale geometric prior that falsifies non-trunk structures via structure-tensor. This is cheap compared to depth, flow or class-supervision methods to create pseudo-labels. These are then used for downstream tasks which we evaluate against other unsupervised baselines. ProGuT achieves a Panoptic Quality (PQ) of $65.2$ on Our-forest dataset (\textbf{$2.6\times$ improvement over the initial pseudo-label quality}) and reaches \textbf{$65.9$ mIoU on Freiburg Forest}, \textbf{outperforming unsupervised baselines like PiCIE ($45.3$ IoU) and STEGO ($57.6$ IoU)}. Additionally, ProGuT outperforms existing unsupervised methods for class-agnostic trunk instance benchmark.

%
% The pseudo-labels train a standard panoptic model to PQ=65.2 on RPTU-Forest (2.6× over the pseudo-labels), and the same pipeline reaches 65.9 mIoU on Freiburg Forest, exceeding STEGO (57.6) and PiCIE (45.3); on a class-agnostic trunk-instance benchmark it substantially outperforms unsupervised baselines.
%
%We present ProGUT, an annotation-free panoptic segmentation pipeline for forest scenes. ProGUT generates pseudo-labels without manual masks by combining CLIP-based semantic clustering with a multi-scale geometric prior that detects tree trunks, followed by downstream panoptic training. Unlike conventional annotation-heavy forest segmentation pipelines, ProGUT uses only unlabeled images and produces both semantic and instance-level pseudo-labels. Experiments on RPTU-Forest and FinnWoodlands show that the generated labels are sufficient to train a downstream panoptic model.
\par
\endgroup

%% file: sec/1_intro.tex
\section{Introduction}
\label{sec:introduction}

When an Autonomous Ground Robot (AGR) enters a forest, it has to answer 2 questions simultaneously: \textit{Where can it drive ?} and \textit{What does it see around itself ?} These questions can be categorized as Semantic or Instance in nature. Panoptic Segmentation addresses both under a unified representation, hence making it an ideal approach for forest robotics. However, deploying panoptic models in forests comes with its own challenges wherein the \textit{"common workflow"} of \textit{train a model on large annotated dataset} breaks down. \\ 
We argue that this breakdown is significantly due to 2 reasons. \textbf{Firstly}, Forests scenes violate every assumption that common instance segmentation methodologies rely on. The most common objects (i.e. tree trunks) are non-salient, often ambiguous, occluded and do not move independently of the scene. The structure breaks the anchor priors and NMS heuristics calibrated for common datasets. \textbf{Secondly}, manual annotations do not scale. Although annotating manually is not the issue, the preciseness and consistency of it is. The process is highly user/task dependent, and often inconsistent. One of the publicly available panoptic datasets (Finnwoodlands \cite{lagos2023finnwoodlands}) has only 300 annotated images from $5,170$ frames and that too on a coarse level. A densely annotated forest dataset at urban level does not exist and the economics of the domain suggests it is difficult. Altogether, these shortcomings motivate towards a different paradigm. A procedure which can be re-run from scratch, without any manual labels, each time the robots encounters a new forest. This is the methodology \textbf{ProGuT (Prototype Guided Training)} is designed to realize.\\
While semantic categorization is somewhat addressed with methods like STEGO \cite{hamilton2022unsupervised} already producing decent stuff maps on forest data (see Table.~\ref{tab:freiburg} \& Figure.~\ref{fig:semantic_comparison}), however, precise instance segmentation is the bottleneck. Existing unsupervised instance discovery methods such as MaskCut \cite{wang2023cut}, CutLER \cite{wang2023cut}, CuVLER \cite{arica2024cuvler} assume that objects are the most salient partitions of the image. But forests violate this assumption with the most salient partition being canopy vs ground. Tree trunks rarely emerge as separate objects, are ambiguous and occluded; the aforementioned methods fail to correctly segregate them (see Figure.~\ref{fig:canatree}). We interpret that this failure is structural and no amount of scale or self-training would repair a broken prior. \\
Instead, with ProGuT, we argue and show that the answer to this is \textbf{\textit{local-image geometry}}. Regardless of species, season, lighting, or camera parameters, a tree trunk projects onto the image plane as a vertically elongated structure whose intensity gradients run predominantly horizontal, across its boundary. This property is the geometric consequence of the trunk's physical shape. At finer scales, a branch might share this property but not at coarser scales. A genuine trunk, as a consequence, is vertically coherent at every scale simultaneously.\textbf{ This finding that trunks can be \emph{falsified} rather than \emph{detected} is the foundation of our approach.} \\
ProGuT is designed as a data engine that produces panoptic pseudo-labels for forest scenes. All components are retrained from scratch on the target environment's images, requiring \textbf{no per-image} masks, only a small labeled reference set used for clusters-to-class assignment. We perform extensive evaluation across three segmentation tasks, 4 datasets and show that the model generalizes well-beyond noisy pseudo-labels. Altogether, our contributions are as follows :
\begin{itemize}
	[topsep=-5px,partopsep=0px]
	\item A multi-scale geometric falsification prior based on the structure tensor that separates trunk instances in dense forest without depth, or motion cues.
	\item ProGuT, a data engine for forest Panoptic Segmentation requiring no per-image training masks.
	\item The quantitative and qualitative demonstration that unsupervised instance discovery methods fail categorically on forest trunk instances, and that a geometry-based prior resolves this failure.
	\item comprehensive evaluation across three tasks and four datasets. 
\end{itemize}

%% file: sec/related_works.tex
\section{Related Work}
\label{sec:related_work}

\subsection{Forest Scene Understanding: Datasets}
\label{sec:forest_datasets}

On one-hand, the task of environment perception in unstructured off-road environments has been largely driven by datasets like RUGD \cite{wigness2019rugd}, RELLIS-3D \cite{jiang2021rellis}, GOOSE \cite{mortimer2024goose}, TGOD \cite{jiang2025go}, Freiburg Forest \cite{valada16iser} and Wildscenes \cite{vidanapathirana2025wildscenes} which are oriented towards traversability (navigable vs non-navigable terrain). These are mostly used for semantic segmentation. On the other hand, RGB-only instance-level forest perception datasets and methods are mostly supervised. Most approaches base their performance on SYNTH43K \cite{grondin2022tree} and CanaTree100 \cite{grondin2022tree} for supervised trunk detection, segmentation and keypoints estimation. Multi-modal datasets such as ForTrunkDet \cite{da2021visible} provides RGB and thermal annotated images for the task of trunk detection and segmentation. Approaches such as TreeLearn \cite{henrich2024treelearn} uses point clouds to segment individual trees from ground-based LiDAR, but needs pre-segmented point clouds for training. FinnWoodlands \cite{lagos2023finnwoodlands}, a forest benchmark which has semantic, instance and panoptic annotations on RGB images, but remains small ($300$ coarsely annotated frames). \\
Here, we observe 2 specific gaps. \textbf{First}, RGB-only tree instance segmentation methods are supervised; a few label-light methods like (\cite{wang2021detectingmappingtreesunstructured}) rely on depth and human review. No mask-free RGB approach exists. \textbf{Second}, the off-road datasets that are \textit{"large enough"} offer only terrain-level semantics, therefore not exactly transferable to a forest setting.  

\subsection{Unsupervised Semantic Segmentation}
\label{sec:rw_semantic}
Annotation-free semantic segmentation was first framed as overclustering. Methods such as IIC~\cite{ji2019invariant} and PiCIE~\cite{cho2021picie} first partitioned the image into many clusters and merged them by enforcing consistency across geometric and photometric transformations. The quality of features was further enhanced by self-supervised vision transformers. For e.g. STEGO \cite{hamilton2022unsupervised} distilled these (DINO \cite{zhang2022dino} patch features) into cluster assignments using contrastive feature correspondences, thereby rivaling early supervised methods in coarse category separation. Parallel approaches have exploited these emerging semantic properties through object-centric slot attention \cite{zadaianchuk2022unsupervised} or by combining saliency maps with feature clustering \cite{sick2024unsupervised}. But on forest/off-road data, these methods transfer poorly by recovering only a few large-homogeneous regions (more details Sec.~\ref{sec:freiburg}). To bypass these representation bottlenecks, methods such as (LSeg \cite{li2022language}, MaskCLIP \cite{dong2023maskclip}, OpenSeg \cite{ghiasi2022scaling}) aligns CLIP features with text for open-vocabulary segmentation. But text prompts assumes a fixed vocabulary, and therefore, we use CLIP features (broader web-scale visual prior) without the text alignment head.

\subsection{Class-Agnostic Instance Discovery}
\label{sec:rw_instance}

On the instance side, approaches such as MaskCut \cite{wang2023cut}, CutLER \cite{wang2023cut} and CuVLER \cite{arica2024cuvler} discover object instances without labels. MaskCut applies normalized cuts \cite{868688} to DINO patch affinities to isolate object-like features. CutLER then bootstraps these masks into a self-trained detector, whereas, CuVLER augments it with additional self-supervised features. All these methods perform well on object-centric benchmarks because their pre-training is dominated by salient objects. Their core assumption being \textbf{\textit{an object is whatever which stands out from its background}}. In forestry, this assumption breaks where the most salient partition is the canopy vs ground and as a consequence, trunks rarely become distinct. Because of this reason, these methods return large blobs rather than individual trees. \textbf{\emph{Hence, we say, that appearance saliency is a wrong prior for forestry environments and therefore, self-training a detector on saliency-based pseudo-masks cannot repair it.} }

\subsection{Unsupervised and Depth-Guided Panoptic Segmentation}
\label{sec:rw_panoptic}

MaskFormer \cite{cheng2021maskformer} unified mask classification with transformer queries towards panoptic segmentation and was further refined by Mask2Former \cite{cheng2022masked} with masked attention, which we adopt in our work. U2Seg \cite{niu2024unsupervised} extended this to an unsupervised regime by pairing unsupervised semantic clustering with MaskCut-based instance discovery (thereby inheriting the same saliency assumption). CUPS \cite{hahn2025scene} is the closest prior work to motivate us: it replaces appearances saliency with metric depth, and identifies things as above-ground, mutually separated clusters. However, their own ablation conveys that depth is the load-bearing component, which makes CUPS inapplicable where depth data is absent. Across all the lineages, the assumption that, \textit{\textbf{instances are separable by appearance, or motion}} prevails which tree trunks don't satisfy. 

\subsection{Foundation Models and Classical Priors}
\label{sec:rw_foundations}

On a different paradigm, prompt-based segmentation methods such as SAM \cite{kirillov2023segment}, SAM2 \cite{ravi2025sam} produce high-quality masks from sparse point or box prompts. Instead of using SAM2 as an object discoverer, we use it as a mask generator driven by geometrically derived point prompts. These points come from structure tensor (\cite{forstner1987fast, bigun1987optimal}), a classical second-moment descriptor of local gradient orientation and anisotropy underlying the Harris corner detector \cite{harris1988combined}, optical flow and coherence-enhancing diffusion \cite{weickert1999coherence, weickert1999coherence_color}. Our contribution is not the tensor itself, but its utilization as a multi-scale falsification filter ($\therefore$ a structure which is vertically non-coherent at multiple scales is rejected as a non-trunk candidate). \\
To our knowledge, this geometrical-falsification prior, together with promptable segmentation to separate dense repeated instances, has not been applied to trunk instance segmentation.

%% file: sec/methodology.tex
\section{Methodology}
\label{methodology}

\subsection{ProGuT - Overview}
\label{sec:progut_overview}
ProGuT is a 5-stage pipeline that produces panoptic pseudo-labels for forest scenes and trains a downstream segmentation model from them. \textbf{\textit{First}}, dense patch features from a pretrained CLIP encoder are clustered with K-means to discover the dataset's dominant visual modes without supervision (Sec.~\ref{sec:clip_clustering}). \textbf{\textit{Second}}, a lightweight UNet refines the coarse patch-level cluster maps into dense per-pixel cluster predictions (Sec.~\ref{sec:unet_refinement}). \textbf{\textit{Third}}, the clusters are mapped to semantic classes through a one-time majority overlap voting procedure (Sec.~\ref{sec:cluster_mapping}) and a hybrid CRF then sharpens stuff-cluster boundaries while preserving trunk geometry, after which $\phi$ produces the n-class semantic map (Subsec.~\ref{subsec:hybrid_crf}). \textbf{\textit{Fourth}}, trunk instances are recovered by a multi-scale geometric prior (Subsec.~\ref{sec:multiscale}) that falsifies non-trunk structures via the structure tensor (Subsec.~\ref{sec:structure_tensor}), localized by column-wise peak detection (Subsec.~\ref{sec:peak_detection}), and segmented by geometry-guided SAM2 prompting (Subsec.~\ref{sec:sam2_prompting}). \textbf{\textit{Fifth}}, the semantic predictions and trunk instances are then assembled into COCO-format panoptic pseudo-labels (Sec.~\ref{sec:panoptic_assembly}).

\subsection{Target-Domain Feature Clustering}
\label{sec:clip_clustering}

We start with feature extraction on the target dataset. The choice of a representation model depends on the environment and specific task. While self-supervised models like DINO are at par with distinct objects, Vision-Language Models (VLMs) like CLIP excel in unstructured natural domains since it captures abstract, universal semantic concepts for cleaner clustering \cite{liu2025data}. We therefore, adopt CLIP ViT-B/16 patch features which cluster more cleanly into forest relevant semantic-groups (path, leaves, trunk \& sky). \\
Once the dense patch features are extracted (at an input resolution of $768\times768$), the encoder produces a $48\times48$ grid of $768$-dimensional patch embeddings, each summarizing a $16\times16$ pixel region. Stacking the grid produces a feature matrix of $\mathbf{F} \in \mathbb{R}^{N \times d}$ per image, with $N = 2304$ patches and $d = 768$\footnote{We find these settings to be suited for all the datasets we evaluated}.

Once the patch features are extracted, we perform global clustering (once) over patch features of all images in the target dataset. We run K-means with $K = 16$ on the pooled set of patch embeddings $\{\mathbf{f}_i\}_{i=1}^{M}$ to obtain a set of centroids (\textbf{Prototypes}) $\{\boldsymbol{\mu}_k\}_{k=1}^{K}$, minimizing
\begin{equation}
	\label{eq:kmeans}
	\sum_{i=1}^{M} \min_{k \in \{1,\dots,K\}}
	\left\lVert \mathbf{f}_i - \boldsymbol{\mu}_k \right\rVert_2^2 
\end{equation}
Clustering globally ensures that the cluster indices are dataset consistent. Each patch is then assigned to its nearest centroid by cosine similarity,
\begin{equation}
	\label{eq:assignment}
	c_i = \arg\max_{k \in \{1,\dots,K\}}
	\frac{\mathbf{f}_i^{\top} \boldsymbol{\mu}_k}
	{\lVert \mathbf{f}_i \rVert_2 \, \lVert \boldsymbol{\mu}_k \rVert_2} ,
\end{equation}
thereby producing $48\times48$ cluster map (\textbf{Prototype labels}) per image.

\subsection{Dense cluster refinement with UNet}
\label{sec:unet_refinement}

The prototype labels obtained are at $48\times48$ patch resolution, far too coarse for any downstream tasks. A $30$ pixels trunk in a $1280\times720$ image may span fewer than 2 patches, and bilinear upsampling of the map may produce blocky, inaccurate boundaries. We bridge this resolution gap with a lightweight UNet trained to predict per-pixel cluster labels at native resolution, supervised by the upsampled patch-level cluster map. We supervise the UNet with a relaxed target that softens the one-hot cluster label with a spatial Gaussian $G_\sigma$ and re-normalizing,
\begin{equation}
	\label{eq:relaxed_target}
	\tilde{\mathbf{t}}_p =
	\frac{\big(G_\sigma * \mathbf{1}[y]\big)_p}
	{\big\lVert (G_\sigma * \mathbf{1}[y])_p \big\rVert_1} ,
\end{equation}
and apply cross-entropy against it,
\begin{equation}
	\label{eq:relaxed_ce}
	\mathcal{L}_{\mathrm{rce}} =
	-\frac{1}{|\Omega|} \sum_{p \in \Omega}
	\sum_{k}
	\tilde{t}_{p,k} \, \log \hat{q}_{p,k} ,
\end{equation}
where $\Omega$ is the set of pixels and $\mathcal{k}$ the indices cluster. This reduces the penalty for predictions matching neighboring patches.

\begin{figure}[!htp]
	\centering
	\includegraphics[width=\linewidth]{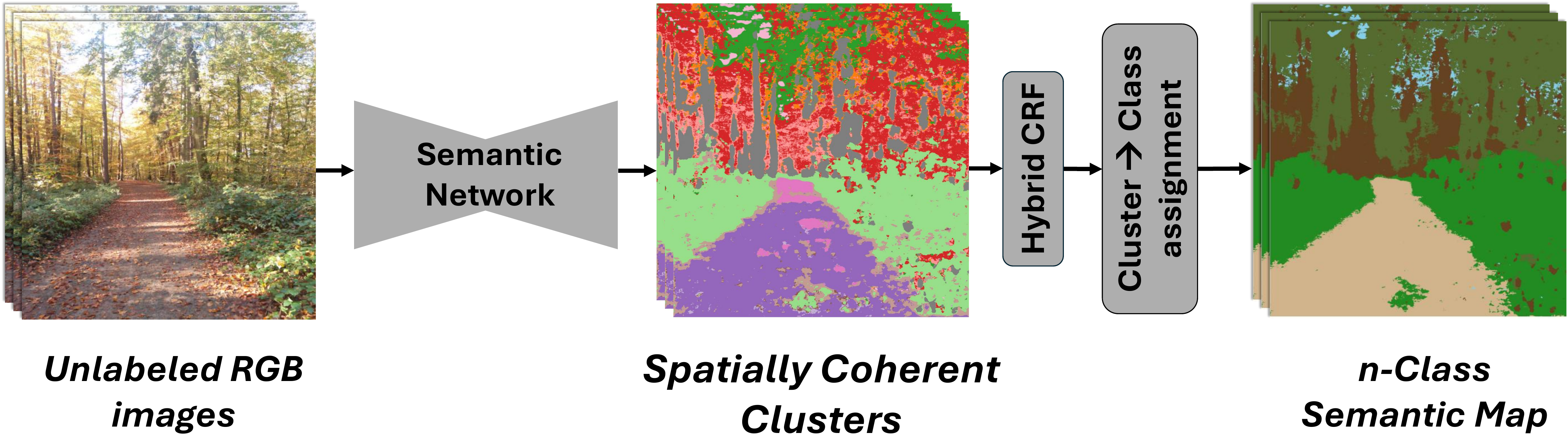}
	\caption{Output of semantic pipeline (stages 1-3): images are clustered, refined with UNet, and mapped to \textit{n}-classes, thereby producing \textit{n}-class semantic map.}
	\label{fig:stage_1_cropped}
\end{figure}

The UNet is trained on all images (train and val) of the target dataset along-with their prototype labels (full details in supplementary). Following standard practices of unsupervised segmentation \cite{hamilton2022unsupervised, hahn2025scene}, the training is transductive (i.e. the model sees the evaluation images but never their labels). The output is dense per-pixel cluster map (\textbf{\textit{Spatially Coherent Clusters}}) (as shown in Figure.~\ref{fig:stage_1_cropped}). 

%The UNet is trained on all images in the deployment dataset without a
%validation split, using all available unlabeled data to capture the target
%environment's visual diversity. We use a cosine schedule with a low initial
%learning rate ($1\!\times\!10^{-6}$); at inference the UNet runs at native
%resolution, producing a dense per-pixel cluster map.

%\begin{figure}[!htp]
%	\centering
%	\includegraphics[width=\linewidth]{images_pdf/stage_2_cropped.pdf}
%	\caption{svg image}
%\end{figure}

%\paragraph{Hybrid CRF.}
%UNet predictions remain spatially imprecise at fine boundaries. DenseCRF~\cite{}
%is the standard tool for sharpening semantic boundaries: its bilateral term
%encourages spatially proximate pixels of similar color to share a label,
%snapping boundaries to high-contrast edges. This smoothing assumption, however,
%is actively harmful for trunks. Trunk boundaries are often low-contrast---a
%grey trunk against a grey-brown background---and thin, so DenseCRF erodes trunk
%regions that are poorly supported by bilateral evidence rather than sharpening
%them. We therefore apply a \emph{hybrid} strategy: DenseCRF refines the stuff
%classes, while the trunk class bypasses CRF entirely. The raw UNet trunk
%prediction is preserved and passed directly to the geometric prior and SAM2
%prompting stages (Sec.~\ref{sec:geometric_prior}--\ref{sec:sam2_prompting}),
%where boundary estimation is handled by SAM2.

\subsection{Cluster-to-Class assignment}
\label{sec:cluster_mapping}
The spatially coherent clusters obtained after UNet training are semantically meaningless. In order to assign a semantic class to clusters, we perform majority-voting. When there is no GT present, a small annotation set ($5$-$10$ images) is created manually (in panoptic format). When GT is present, the training annotations are used for majority voting. In all cases, the labels only inform the mapping-$\phi$, never the training.

Let $\mathcal{C}$ denote the labeled set, $\mathcal{P}_k$ the pixels assigned to cluster $k$ (after upsampling the cluster map to image resolution), and $G_s$
the pixels labeled class $s$. Each cluster is assigned to the class of maximal pixel overlap,
\begin{equation}
	\label{eq:majority_vote}
	\phi(k) = \arg\max_{s}
	\sum_{p \in \mathcal{C}}
	\left\lvert \mathcal{P}_k(p) \cap G_s(p) \right\rvert ,
\end{equation}
thereby yielding a lookup $\phi : \{1,\dots,K\} \rightarrow \mathcal{S}$.

However, trunks are relatively thin compared to the $16$-pixel patch, small image patches often contain a majority of background pixels, preventing any cluster from winning a plurality under Eq. \ref{eq:majority_vote}. To address this, we apply a \emph{column-rescue}: if no cluster maps to the trunk class, the cluster of largest absolute trunk-pixel count is reassigned to it,
\begin{equation}
	\label{eq:trunk_rescue}
	\phi(k^{\star}) = \texttt{trunk}, \quad
	k^{\star} = \arg\max_{k}
	\sum_{p \in \mathcal{C}}
	\left\lvert \mathcal{P}_k(p) \cap G_{\texttt{trunk}}(p) \right\rvert .
\end{equation}

%The $K$ clusters from Sec.~\ref{sec:clip_clustering} carry no semantic meaning.
%Assigning each to a class is the only stage of ProGUT requiring human input, and
%it is deliberately lightweight: a small set of coarsely labeled calibration
%images suffices, providing region labels rather than per-instance masks, once
%per deployment. Let $\mathcal{C}$ be the calibration set, $\mathcal{P}_k$ the
%pixels assigned to cluster $k$ (after upsampling the cluster map to image
%resolution), and $G_s$ the pixels labeled class $s$. We assign each cluster to
%the class it most overlaps,
%\begin{equation}
%	\label{eq:majority_vote}
%	\phi(k) = \arg\max_{s}
%	\sum_{p \in \mathcal{C}}
%	\left\lvert \mathcal{P}_k(p) \cap G_s(p) \right\rvert ,
%\end{equation}
%defining a lookup $\phi : \{1,\dots,K\} \rightarrow \mathcal{S}$.
%
%Trunks need special handling: thin relative to the $16$-pixel patch, a
%boundary-straddling patch holds more background than trunk and is voted to a
%background class, so no cluster may win a plurality of trunk pixels. We add a
%\emph{column-rescue} pass---if no cluster maps to trunk, the cluster with the
%largest absolute trunk-pixel count is reassigned to it,
%\begin{equation}
%	\label{eq:trunk_rescue}
%	\phi(k^{\star}) = \texttt{trunk}, \quad
%	k^{\star} = \arg\max_{k}
%	\sum_{p \in \mathcal{C}}
%	\left\lvert \mathcal{P}_k(p) \cap G_{\texttt{trunk}}(p) \right\rvert .
%\end{equation}
%The mapping $\phi$ is fixed after calibration and applied to all images without
%further input.

\subsubsection{Hybrid CRF}
\label{subsec:hybrid_crf}
UNet predictions remain spatially imprecise at fine boundaries. We apply DenseCRF to all cluster predictions, after which pixels belonging to the trunk cluster are restored from the un-smoothed prediction. The trunk cluster is identified directly from the mapping $\phi$ (Sec.~\ref{sec:cluster_mapping}), so this restoration operates in cluster space, before the final cluster-to-class assignment. Towards the end, we get \textbf{\textit{n-class semantic map}} (as shown in Figure.~\ref{fig:stage_1_cropped})

\subsection{Multiscale Geometric Prior for Trunk Discovery}
\label{sec:geometric_prior}

As previously stated in Sec.~\ref{sec:introduction}, appearance, depth and motion are not enough to separate trunk instances. Rather than asking \textit{what does a trunk look like ?}, we ask \textit{what it cannot fail to be ?} \textit{\textbf{A trunk should be vertically coherent at every scale.}} Any candidate that fails this condition gets rejected. We term it as \textbf{Geometric Falsification}.  

%Appearance, depth, and motion---the three cues standard instance segmentation
%relies on---are all unreliable for separating trunks in dense forest
%(Sec.~\ref{sec:related_work}). We instead exploit a signal that holds
%regardless of species, season, or lighting: a trunk projects onto the image as
%a vertically elongated structure whose intensity gradients run predominantly
%horizontal, across its boundary. This is a geometric consequence of the trunk's
%shape, not an appearance property, and it is the basis of our instance
%separation mechanism. Rather than asking what a trunk looks like, we ask what a
%trunk cannot fail to be---vertically elongated at every scale at which it is
%visible---and discard candidates that fail this test. We refer to this as
%\emph{geometric falsification}.

\subsubsection{Structure Tensor Coherence and Orientation}
\label{sec:structure_tensor}

In an image, the local features can be summarized by calculating the gradient orientation and edge-strength across a neighborhood. This is called Structure Tensor (\cite{forstner1987fast, bigun1987optimal}). For an image $I$ with spatial gradients $I_x, I_y$, the structure tensor at scale $\sigma$ is the Gaussian-smoothed outer product of the gradient,

\begin{equation}
	\label{eq:structure_tensor}
	J_\sigma = G_\sigma *
	\begin{pmatrix}
		I_x^2 & I_x I_y \\
		I_x I_y & I_y^2
	\end{pmatrix} ,
\end{equation}

where $G_\sigma$ is a Gaussian kernel of standard deviation $\sigma$ and $*$ signifies convolution. The eigenvalues $\lambda_1 \geq \lambda_2 \geq 0$ of $J_\sigma$ characterize the local structure: $\lambda_1 \gg \lambda_2$ indicates a strongly oriented edge-like region, $\lambda_1 \approx \lambda_2$ indicates an isotropic or corner region, and both near zero indicates a flat region. From the eigenvalues we form the coherence,

\begin{equation}
	\label{eq:coherence}
	C = \left( \frac{\lambda_1 - \lambda_2}{\lambda_1 + \lambda_2} \right)
	\in [0, 1] ,
\end{equation}

which is near $1$ for strongly oriented structure. With $\theta$ the edge orientation (perpendicular to the dominant gradient), a pixel is trunk-coherent at scale
$\sigma$ when both strongly oriented and near-vertical, 

\begin{equation}
	\label{eq:coherence_mask}
	m_\sigma(p) =
	\mathbf{1}\!\left[\, C_\sigma(p) \geq \tau_C \;\wedge\;
	\theta_\sigma(p) \geq \tau_\theta \,\right],
\end{equation}
with $\tau_\theta = 55^{\circ}$.

\begin{figure}[!htp]
	\centering
	\includegraphics[width=\linewidth]{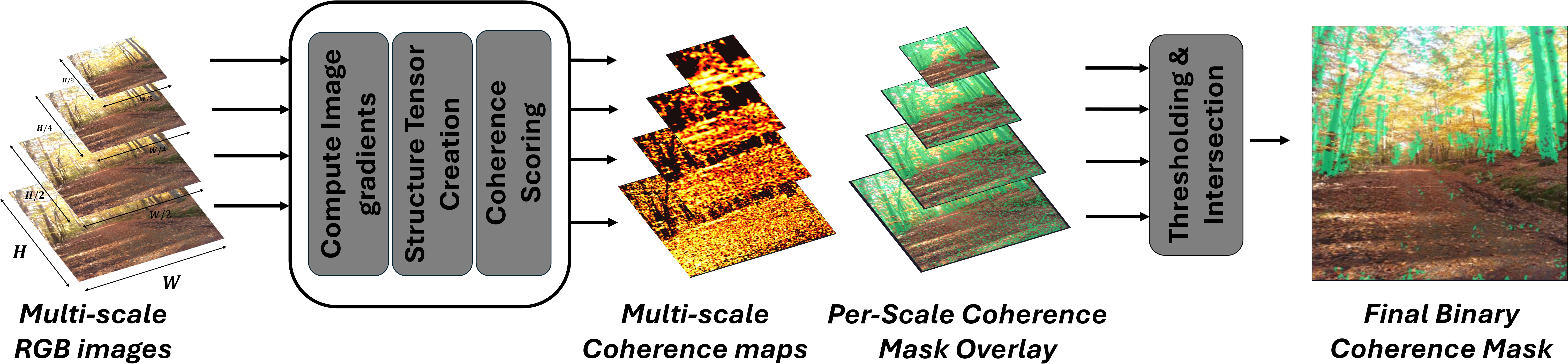}
	\caption{Multi-scale coherence:  the coherence map is produced by a structure tensor at each scale, followed by thresholding and intersecting across scales to retain structure that is vertically coherent at every scale.}
	\label{fig:multiscale}
\end{figure}

\subsubsection{Multiscale Intersection for Geometric Falsification}
\label{sec:multiscale}

We observe that at a single scale, the coherence pattern is mixed (noise with a pattern for trees). At fine scales, branches, bark and shadow edges all produce high-coherence with near-vertical orientation; at coarse scale, only large structures prevail. \textbf{\textit{A genuine trunk should be coherent at \emph{every} scale, and false positives should fail at some scale.}} We, therefore, compute coherence mask at 4 scales $\sigma \in \{\sigma_1, \sigma_2, \sigma_3, \sigma_4\}$, corresponding to effective spatial supports of approximately $8$, $16$, $32$, and $64$ pixels, and retain only the pixels coherent at all four,

\begin{equation}
	\label{eq:intersection}
	M(p) = \bigwedge_{\sigma \in \{\sigma_1,\dots,\sigma_4\}} m_\sigma(p) .
\end{equation}

This intersection showcases falsification directly and yields a sparse, high-precision mask $M$ (as seen in Figure.~\ref{fig:multiscale}). 

\subsubsection{Column-Wise Peak Detection for Trunk Candidate Localization}
\label{sec:peak_detection}

The coherence mask $M$ marks trunk pixels but does not separate them. Since trunks are "somewhat" vertical, we project $M$ onto image columns, 
\begin{equation}
	\label{eq:column_projection}
	P(x) = \frac{1}{H} \sum_{y=1}^{H} M(x, y) ,
\end{equation}
where $H$ is the image height. \\
The signal $P(x)$ is a one-dimensional representation of trunk presence across the image width: columns dominated by a trunk produce high values, and vice-versa whereas the peaks of $P$ correspond to trunk centers. However, sometimes, the column responses of closely spaced trunks can be merged into a single broad peak at coarse smoothing. We address this using a \textit{hierarchical peak search }wherein the coarse peaks at $\sigma_{\mathrm{coarse}} = 8$ are detected and split into sub-peaks at $\sigma_{\mathrm{fine}}=6$, thereby separating touching trunks. Each peak $x_j$ yields a seed $(x_j, y_j)$ at its strongest-coherence row.

\subsubsection{Geometry-Guided SAM2 Prompting}
\label{sec:sam2_prompting}

\begin{figure}[!htp]
	\centering
	\includegraphics[width=\linewidth]{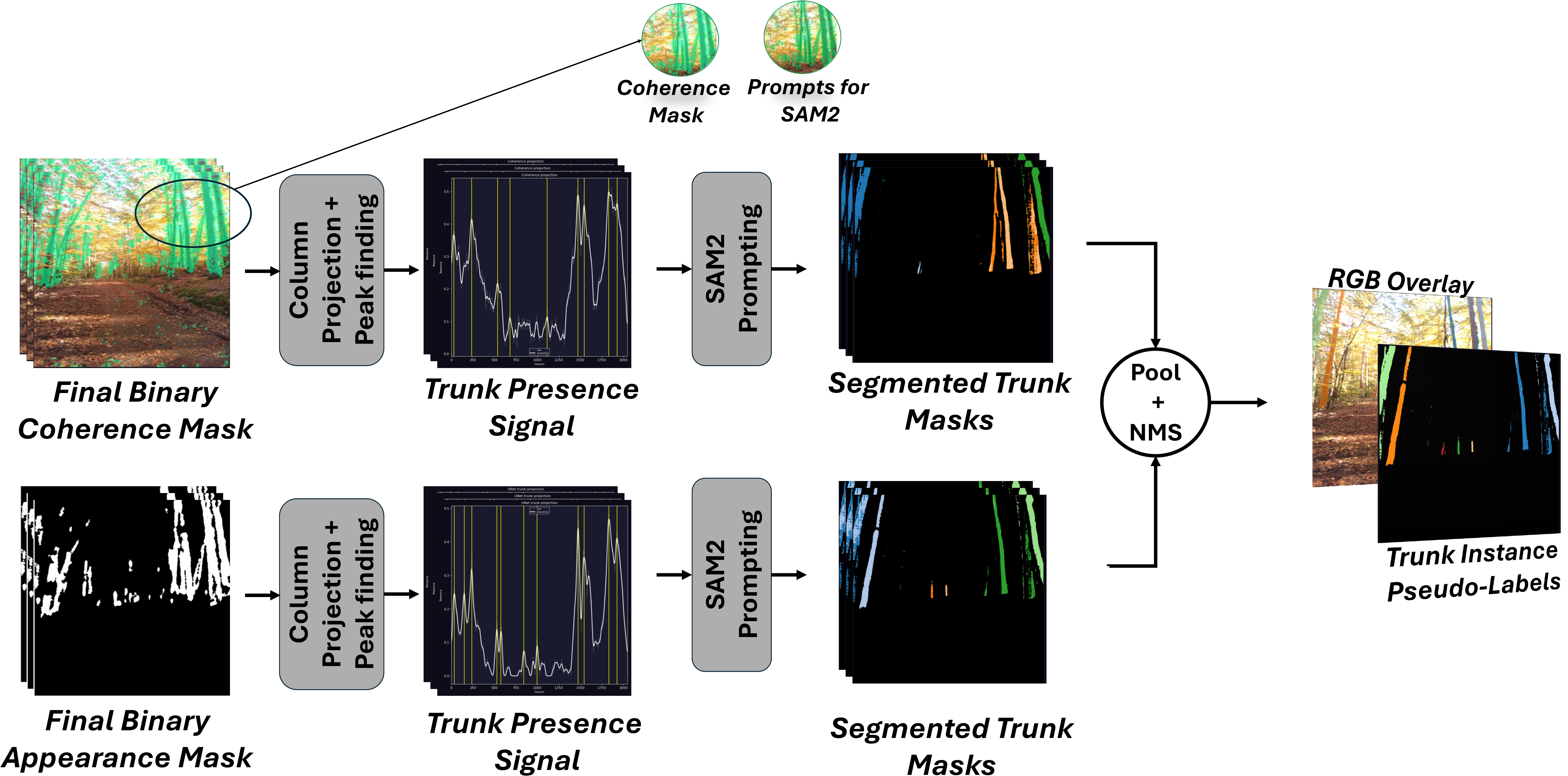}
	\caption{\textbf{Dual-prior trunk instancing.} The coherence and appearance priors are projected to per-column trunk-presence signals; the peaks of which are used for SAM2 prompting (+ve peak center, -ve edges) that segment individual trunks. Trunks from both the priors are then pooled and de-duplicated by NMS into final trunk instance pseudo-labels. }
	\label{fig:sam2_dual}
\end{figure}

In the next step, each candidate peak $x_j$ acts as a prompt for SAM2. A single positive point is placed at the column center, at the row of the strongest coherence response within that column; 4 negative points at the image edges suppress masks that leak into background. SAM2 returns its highest-scoring candidate , which is discarded if its overlap with the prior mask falls below a threshold. \\
Trunk candidates emerge from 2 priors i.e. \textbf{Geometric} and \textbf{Appearance}, each prompting SAM2 independently (see Figure.~\ref{fig:sam2_dual}). The \emph{Geometric} pass takes peak from multiscale coherence mask $M$ whereas, the $\emph{appearance}$ pass takes peaks from the raw UNet trunk predictions. The 2 trunk candidates obtained are then pooled and de-duplicated by NMS.

%Finally, the geometric prior is grounded by an appearance veto. The structure
%tensor detects vertical elongated structures but does not know they are trees;
%in scenes containing other vertical structures it would fire on them as well.
%We therefore suppress any candidate mask whose overlap with the UNet trunk
%prediction (Sec.~\ref{sec:unet_refinement}) falls below a minimum threshold.
%Geometry is reliable where appearance fails---bare winter trunks, low
%contrast---while the appearance veto supplies semantic grounding where geometry
%is ambiguous; the two priors are complementary.

\subsection{Panoptic pseudo-label Assembly}
\label{sec:panoptic_assembly}
The final trunk instances and stuff predictions are composited into a COCO-format panoptic label by priority. Stuff classes act as background and the trunk instances are painted over this background, overriding the stuff labels at those pixels. Trunk pixels not claimed by any instance fall back to the dominant stuff class. These are the final \textbf{\textit{pseudo-labels}} which are then used for downstream tasks.

\begin{figure}[!htp]
	\centering
	\includegraphics[width=\linewidth]{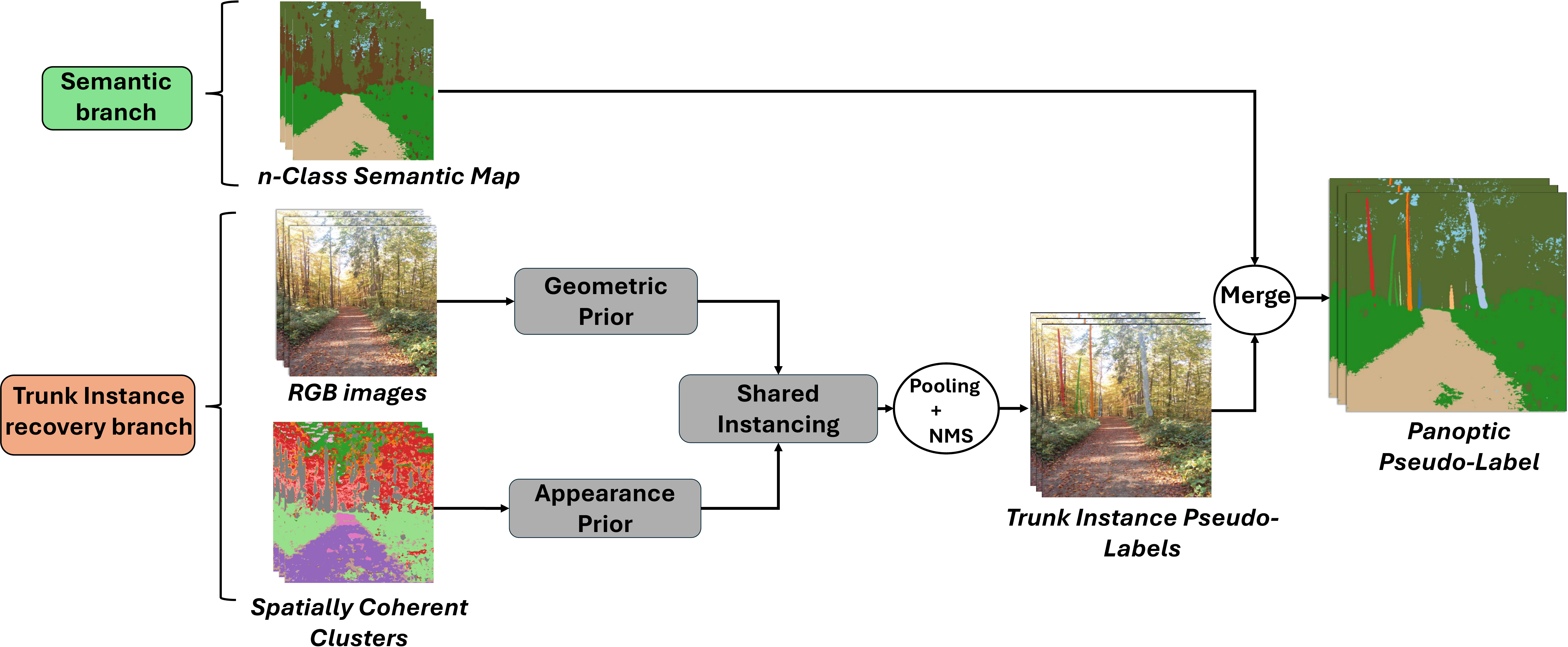}
	\caption{Trunk-instance branch: geometric and appearance prior feed into a shared instancing (column peaks + SAM2), pooled and de-duplicated with NMS into trunk instances, then merged with \textit{n}-class semantic map to produce panoptic pseudo-label.}
\end{figure}

%% file: sec/experimentations.tex
\subsection{Datasets and Evaluation Protocol}
\label{sec:datasets}

We evaluate ProGuT on 4 forest datasets spanning 3 tasks: Panoptic segmentation on \textbf{Our forest} (\ref{our-forest}) and \textbf{Finnwoodlands}, semantic segmentation on \textbf{Freiburg Forest}, and zero-shot instance segmentation on \textbf{CanaTree100}. On Finnwoodlands we remap the $300$ annotated images into 4 coarse classes (trunk, ground, sky, other). For Freiburg Forest, we report the standard 4-class variant (sky, trail, grass and vegetation) whereas evaluation on CanaTree100 (100 images) is purely zero-shot. \textit{Across all datasets, the labels are only used for cluster-to-class assignment and evaluation but never during training.} We report PQ, SQ and RQ (decomposed into Things and Stuff) for panoptic segmentation, mIoU for semantic segmentation and mask AP/AP50/AP75 for instance segmentation.

\paragraph{Our forest}
\label{our-forest} 
Our primary evaluation dataset comprises of ${\sim}3{,}550$ images collected with a multispectral camera ($3456\times4608$, resized to $2048\times2048$) in a forest (autumn). We evaluate using 5-class ontology (Tree trunk as thing; grass shrubs, path, leaves and sky as stuff). 21 images were manually annotated in COCO panoptic format for evaluation (15) and cluster-to-class assignment (6).

\section{Experiments}
\subsection{Semantic Segmentation}
\label{sec:freiburg}

We test the competitiveness of ProGuT's semantic stage on freiburg forest dataset and compare against two unsupervised semantic segmentation methods PiCIE \cite{cho2021picie} and STEGO \cite{hamilton2022unsupervised} and report supervised methods for reference. We evaluate on 4-class variant (sky, vegetation, grass and trail). 

%To test whether ProGUT's semantic stage is competitive independent of the
%panoptic pipeline built on top of it, we evaluate on Freiburg Forest, a pure
%semantic segmentation benchmark. We compare against two annotation-free
%semantic methods---STEGO~\cite{stego} and PiCIE~\cite{picie}---and report
%supervised methods for reference. Following standard protocol on this benchmark,
%we evaluate the four-class variant (sky, trail, grass, vegetation).

\begin{figure*}[!htp]
	\centering
	\includegraphics[width=\linewidth]{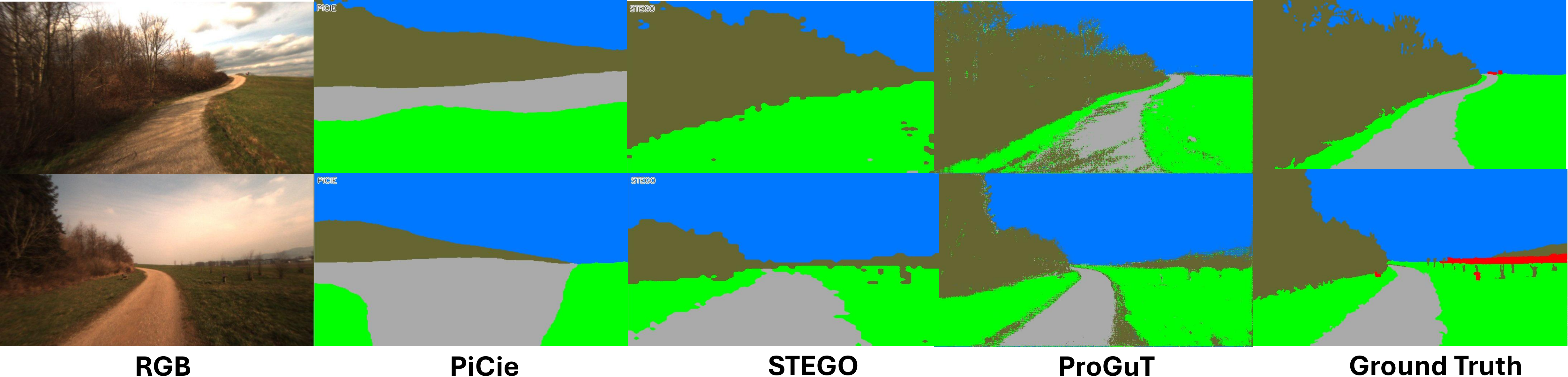}
	\caption{Qualitative comparison of PiCIE \cite{cho2021picie}, STEGO \cite{hamilton2022unsupervised} and ProGuT on Freiburg forest dataset \cite{valada16iser}.}
	\label{fig:semantic_comparison}
\end{figure*}

For ProGuT, the semantic pseudo-labels are the direct output of the UNet stage (Sec.~\ref{sec:unet_refinement}), and does not include SAM2 instancing. We report 2 configurations i.e. ProGuT-UNet is the UNet cluster prediction (mapped through $\phi$) directly. ProGuT-DeepLabV3 trains DeepLabV3 (ResNet-50, ImageNet-pretrained, $50$ epochs, batch size $8$) on training-split pseudo-labels; the checkpoint is selected on a held-out validation split of the training set and evaluated once on the $136$ test images. The test set is never used for model selection. 

%For ProGUT, the semantic pseudo-labels are the direct output of the UNet stage
%(Sec.~\ref{sec:unet_refinement}), with no SAM2 instancing, since Freiburg Forest
%has no thing classes. We report two configurations.
%\textbf{ProGUT-UNet} is the UNet prediction itself, evaluated directly.
%\textbf{ProGUT+DeepLabV3} trains a DeepLabV3 model with a ResNet-50 backbone
%(ImageNet-pretrained, $50$ epochs, batch size $8$) on ProGUT pseudo-labels
%generated from the $366$ annotated images, and evaluates on the $136$ test
%images. We use DeepLabV3 rather than Mask2Former here deliberately: Freiburg
%requires only dense per-pixel prediction, and Mask2Former's object-query
%machinery adds complexity with no benefit on a task with no instances.

\begin{table}[t]
	\centering
	\caption{Semantic segmentation on freiburg forest (4-class mIoU). All mask-free methods use no manual training masks. $\dagger$: supervised, shown for reference.}
	\label{tab:freiburg}
	\setlength{\tabcolsep}{4pt}
	\begin{tabular}{lccccc}
		\toprule
		\textbf{Method} & \textbf{Sky} & \textbf{Trail} & \textbf{Grass} & \textbf{Veg}. & \textbf{mIoU} \\
		\midrule
		PiCIE \cite{cho2021picie}             & 70.41 & 17.69 & 46.68 & 46.24 & 45.25 \\
		STEGO \cite{hamilton2022unsupervised}              & 73.50 & 31.20 & 59.05 & 66.53 & 57.57 \\
		ProGuT-UNet        & 79.23 & 32.00 & 54.96 & 66.54 & 58.18 \\
		\textbf{ProGuT+DeepLabV3}   & \textbf{85.29} & \textbf{38.88} & \textbf{65.51} & \textbf{73.89} & \textbf{65.89} \\
		\midrule
		E-Net$\dagger$ \cite{oliveira2020singlestageencoderdecodernetworks}     & -- & -- & -- & -- & 71.40 \\
		SegNet$\dagger$ \cite{valada2017icra}    & -- & -- & -- & -- & 74.81 \\
		FCN8$\dagger$ \cite{valada2017icra}    & -- & -- & -- & -- & 77.46 \\
		DABNet$\dagger$ \cite{edlinger2023terrain}   & -- & -- & -- & -- & 81.50 \\
		DD-Net$\dagger$ \cite{oliveira2020singlestageencoderdecodernetworks}   & 92.90 & 88.90 & 88.50 & 90.70 & 90.20 \\
		\bottomrule
	\end{tabular}
\end{table}

ProGuT-DeepLabV3 (Table.~\ref{tab:freiburg}) achieves a mask-free state-of-the-art (\textbf{SOTA})\textbf{ $65.89$} mIoU, \textbf{surpassing STEGO by $8.3$ points and PiCIE by $20.6$ points}, while ProGuT-UNet alone is competitive with existing methods at \textbf{$58.18$} mIoU. The $7.71$-point increase with ProGuT-DeepLabV3 shows that the downstream model recovers boundary precision and class consistency (Figure.~\ref{fig:semantic_comparison}), which the clustering based UNet cannot reach alone.

\subsection{Zero-Shot Trunk Instance Segmentation}
\label{sec:canatree}

We evaluate the robustness of ProGuT's trunk discovery component with CanaTree100 \cite{grondin2022tree} dataset against 3 unsupervised instance segmentation methods i.e. MaskCut, CutLER and CuVLER. The evaluation protocol is similar; no method uses CanaTree100 training data; all are evaluated zero-shot against GT, averaged over 5-fold protocol. For ProGuT, we evaluate the SAM2 Pseudo-label masks (Sec.~\ref{sec:sam2_prompting}) directly, with no downstream detector training.

%CanaTree100 isolates the instance-separation problem: per-trunk masks with no
%stuff classes. We use it to test ProGUT's trunk discovery directly, against
%three annotation-free instance methods---MaskCut~\cite{maskcut},
%CutLER~\cite{CutLER}, and CuVLER~\cite{cuvler}. No method uses CanaTree100
%training data; all are evaluated zero-shot against the GT, averaged over the
%five-fold protocol. For ProGUT we evaluate the SAM2 pseudo-label masks
%(Sec.~\ref{sec:sam2_prompting}) directly, with no downstream detector---a point
%we return to below.

\begin{table}[t]
	\centering
	\caption{Instance segmentation on CanaTree100 (mean AP over 5-fold CV). No method uses CanaTree100 training data.}
	\label{tab:canatree}
	\begin{tabular}{lccc}
		\toprule
		\textbf{Method} & \textbf{AP} &\textbf{ AP50} & \textbf{AP75} \\
		\midrule
		MaskCut \cite{wang2023cut}                      & 0.00 & 0.00 & 0.00 \\
		CuVLER \cite{arica2024cuvler}                       & 0.22 & 0.25 & 0.25 \\
		CutLER \cite{wang2023cut}                       & 0.53 & 0.82 & 0.52 \\
		\midrule
		ProGuT (coherence-only)       & 13.65 & 28.46 & 12.05 \\
	\textbf{ProGuT (dual-pass) }           & \textbf{16.60} & \textbf{34.20} & \textbf{14.74} \\
		\bottomrule
	\end{tabular}
\end{table}

\begin{figure*}[htp!]
	\centering
	\includegraphics[width=\linewidth]{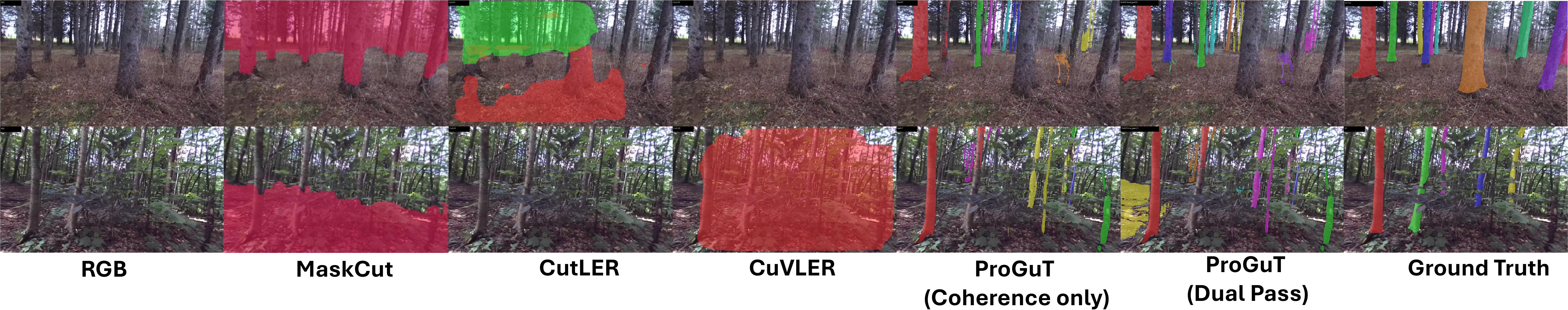}
	\caption{Qualitative evaluation of MaskCut \cite{wang2023cut}, CutLER \cite{wang2023cut}, CuVLER \cite{arica2024cuvler} and ProGuT on CanaTree100 dataset.}
	\label{fig:canatree}
\end{figure*}

We see that all methods collapse on CanaTree100 with AP50 below $1\%$ (Table.~\ref{tab:canatree} and Figure.~\ref{fig:canatree}). As discussed earlier (in Sec.~\ref{sec:related_work}), all the three detect large background regions rather than individual trunks.\footnote{Since normalized cut on self-supervised features finds the most salient partition of the scene,which in a forest is never trunk-versus-background} ProGuT's Dual pass, in contrast, reaches AP50 $=34.20$, a \textbf{$40\times$ }improvement over CutLER, depicting the superiority of geometric falsification over saliency-based methods.  \\
Since CanaTree100 provides only 100 images, we report the results as raw pseudo-labels to avoid performance degradation when training on limited data (E.g. on Mask-RCNN). We thus argue that these $34\%-$AP50 masks provide a strong starting point for cost-effective bootstrapping compared to the supervised upper bound of $69.36$-AP50.

%
%The contrast is categorical. MaskCut, CutLER, and CuVLER all collapse on
%CanaTree100, with AP50 below $1\%$. Inspection confirms the failure mode argued
%in Sec.~\ref{sec:related_work}: all three detect large background
%regions---canopy, ground, sky---rather than individual trunks, because the
%normalized cut on self-supervised features finds the most salient partition of
%the scene, which in a forest is never trunk-versus-background. ProGUT dual-pass
%reaches AP50 $=34.20$, a $40\times$ improvement over CutLER. The geometric
%falsification prior succeeds precisely where saliency-based discovery fails.

%\paragraph{Dataset-dependence of the appearance prior.}
%On CanaTree100, dual-pass (AP $16.60$) outperforms coherence-only (AP $13.65$),
%in contrast to RPTU (Table~\ref{tab:rptu_ablation}), where the appearance prior
%adds nothing. This is consistent rather than contradictory: the UNet appearance
%prior contributes in proportion to how cleanly CLIP clusters isolate trunks,
%which varies by dataset. The trunk-dominant cluster has $25.1\%$ trunk purity on
%CanaTree100, where the appearance prior helps; on RPTU it is marginal; and on
%FinnWoodlands, at $13.5\%$ purity, the prior is uninformative and dual-pass
%reduces to coherence-only (Sec.~\ref{sec:finnwoodlands}). The geometric prior is
%the constant that carries trunk separation across all three; the appearance
%prior is a dataset-dependent bonus.

%\paragraph{Why pseudo-labels, not a downstream model.}
%Unlike the RPTU and Freiburg tracks, we report ProGUT pseudo-labels directly on
%CanaTree100 rather than a downstream detector. The five-fold protocol provides
%only $60$ training images per fold, too few for a Mask R-CNN to generalize from
%noisy pseudo-labels; a downstream model trained on this regime degraded rather
%than improved over the pseudo-labels, so we omit it. The numbers above are
%therefore a lower bound on what ProGUT supports on this dataset---they reflect
%raw pipeline output, not a trained model. The gap to the supervised upper bound
%(AP50 $69.36$) is the cost of zero annotation, and the pseudo-labels are a
%strong starting point for annotation-assisted bootstrapping: correcting
%$34\%$-AP50 masks is far cheaper than annotating from scratch.

\subsection{Panoptic Segmentation}
\label{sec:rptu_main}

Additionally, we evaluate ProGuT's main strength i.e. Panoptic segmentation on Our forest dataset against a held-out set of $15$ manually annotated images. The cluster-to-class mapping was calibrated on a separate set of $6$ annotated images, disjoint from these $15$; the evaluation images are therefore unseen by every stage of the pipeline, including the mapping. Pseudo-labels are generated for all $\sim\!3{,}550$ images and used to train Mask2Former model with Swin-Tiny backbone for $50$k iterations; 

%We evaluate ProGUT on RPTU-Forest against a held-out set of $15$ manually
%annotated images. Pseudo-labels are generated for all $\sim\!3{,}600$ images
%and used to train a Mask2Former model with a Swin-Tiny backbone for $50$k
%iterations; the same $15$ images serve as ground truth for both the
%cluster-to-class calibration and this evaluation, a constraint we return to in
%Sec.~\ref{sec:rptu_ablation}.

\begin{table}[t]
	\centering
	\caption{Panoptic segmentation on Our forest ($15$ held-out GT images). U2Seg \cite{niu2024unsupervised} is the unsupervised panoptic baseline (MaskCut instance discovery +	STEGO semantic clustering, Hungarian-matched to Our forest classes). ProGuT	pseudo-labels are the direct pipeline output; ProGuT + Mask2Former is the
		downstream model trained on them.}
	\label{tab:rptu_main}
	\begin{tabular}{lccc}
		\toprule
		\textbf{Method} & \textbf{PQ} & \textbf{PQ$_{\text{Th}}$} &\textbf{ PQ$_{\text{St}}$} \\
		\midrule
		U2Seg \cite{niu2024unsupervised}               & 3.75  & 0.00  & 11.33 \\
		ProGuT (pseudo-labels) & 25.13 & 17.80 & 43.16 \\
		\textbf{ProGuT + Mask2Former} & \textbf{65.18} & \textbf{29.21} & \textbf{74.17} \\
		\bottomrule
	\end{tabular}
\end{table}

\begin{figure}[!htp]
	\centering
	\includegraphics[width=\linewidth]{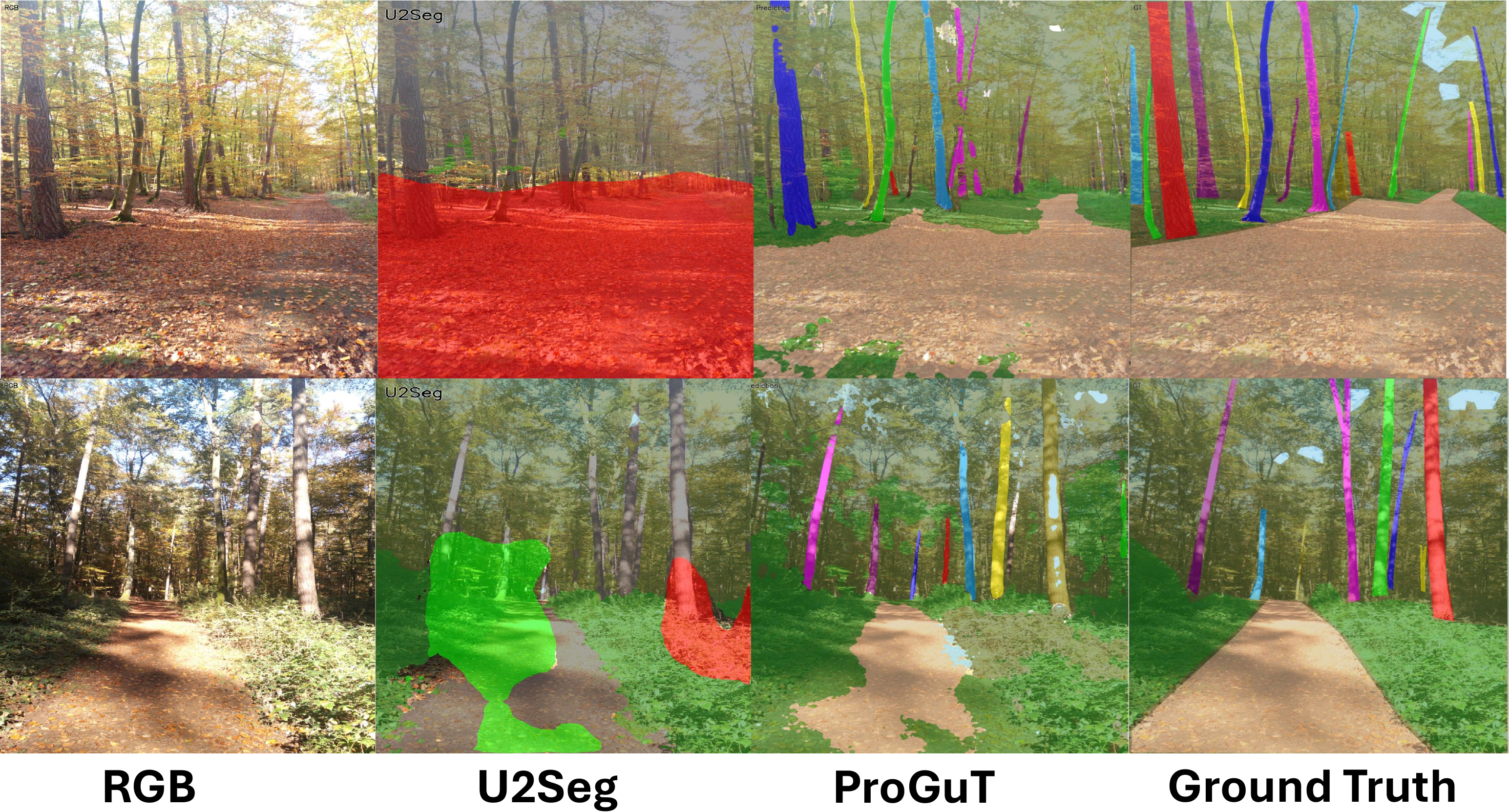}
	\caption{Qualitative results of U2Seg \cite{niu2024unsupervised} and ProGuT on Our forest dataset}
	\label{fig:panoptic_track}
\end{figure}

We report the results of panoptic track with Table~\ref{tab:rptu_main} and Figure.~\ref{fig:panoptic_track}. Mask2Former trained on ProGuT pseudo-labels reaches $65.18$ PQ, a $2.6\times$ improvement over the pseudo-labels it was trained on $25.13$. Similarly, we see that the downstream model generalizes well beyond its noisy training signal. However, we also see that the pseudo-labels themselves, also carry sufficient signal despite their noise. The unsupervised baseline, U2Seg reaches only $3.75$ PQ with zero thing quality. We further report in Table~\ref{tab:rptu_ablation}, the individual components at the pseudo-label stage, on the same $15$ held-out images. 

%Table~\ref{tab:rptu_main} reports the central result of the panoptic track.
%Mask2Former trained on ProGUT pseudo-labels reaches $65.18$ PQ, a
%$2.6\times$ improvement over the pseudo-labels it was trained on
%($25.13$ PQ). This gap is the empirical evidence that the downstream model
%generalizes well beyond its noisy training signal: it fills in trunks the
%pipeline missed, completes fragmented masks, and corrects imprecise stuff
%boundaries. The result also confirms that the pseudo-labels, despite their
%noise, carry sufficient signal---trunk locations and class boundaries are
%present, if imperfectly expressed, and the transformer recovers and regularizes
%them. Stuff segmentation is strong ($74.17$ PQ$_{\text{St}}$); the lower
%thing score ($29.21$ PQ$_{\text{Th}}$) reflects the residual difficulty of
%exact trunk delineation, which we analyze further in
%Sec.~\ref{sec:finnwoodlands} and Sec.~\ref{sec:canatree}.
% VERIFY: SQ/RQ available if you want a wider table --
% baseline All: PQ 65.18 / SQ 77.83 / RQ 82.32
% Things: 29.21 / 70.72 / 41.30 ; Stuff: 74.17 / 79.61 / 92.57

%
%\subsection{Component Ablation on RPTU-Forest}
%\label{sec:rptu_ablation}
%
%Table~\ref{tab:rptu_ablation} isolates the contribution of each pipeline
%component at the pseudo-label stage, evaluated on the same $15$ GT images. Each
%row adds one component to the previous.

\begin{table}[t]
	\centering
	\caption{Pseudo-label ablation on Our forest using $15$ GT images. Each row adds one component over the row above.}
	\label{tab:rptu_ablation}
	\scriptsize
	\setlength{\tabcolsep}{2.2pt}
	\renewcommand{\arraystretch}{1.08}
	\begin{tabularx}{\columnwidth}{@{}>{\raggedright\arraybackslash}Xccccc@{}}
		\toprule
		\textbf{Configuration} & \textbf{PQ} & \textbf{SQ} & \textbf{RQ} & \textbf{PQ$_{\mathrm{Th}}$} & \textbf{PQ$_{\mathrm{St}}$} \\
		\midrule
		Semantic only        & 11.67 & 61.55 & 18.97 & 0.00  & 28.21 \\
		\hspace{0.4em}+ coherence & 22.99 & 64.28 & 35.76 & 16.67 & 36.55 \\
		\hspace{0.4em}+ UNet       & 20.18 & 64.18 & 31.45 & 14.07 & 34.33 \\
		\hspace{0.4em}+ dual-pass  & 20.69 & 63.61 & 32.53 & 15.15 & 34.32 \\
		\hspace{0.4em}+ dual-pass + CRF & 25.13 & 69.53 & 36.15 & 17.80 & 43.16 \\
		\bottomrule
	\end{tabularx}
\end{table}

We summarize the findings:
	\begin{itemize}
		\item Geometric prior is the primary source of Thing detection; coherence-only raises PQ$_{\text{Th}}$ from $0.00$ to $16.67$, while the appearance-based UNet-only prior reaches only $14.07$.
		\item Hybrid-CRF contributes the largest single jump in stuff quality (PQ$_{\text{St}}$ $34.32 \rightarrow 43.16$).
		\item The high performance of ProGuT-Mask2Former shows generalization ability that the pseudo-labels alone cannot reach ($25.13$ vs $65.18$)
	\end{itemize}

%Three findings emerge. First, the geometric prior is the dominant source of
%thing detection: coherence-only raises PQ$_{\text{Th}}$ from $0.00$ to
%$16.67$, while the appearance-based UNet-only prior reaches only $14.07$, and
%combining them (dual-pass) does not improve over coherence-only. On RPTU the
%geometric falsification carries trunk separation; the appearance prior is
%marginal. Second, the hybrid CRF contributes the largest single jump in stuff
%quality (PQ$_{\text{St}}$ $34.32 \rightarrow 43.16$), confirming that CRF
%sharpens stuff boundaries while the trunk class correctly bypasses it. Third,
%the full pipeline pseudo-labels reach only $25.13$ PQ, underscoring that the
%$65.18$ PQ of the trained model (Sec.~\ref{sec:rptu_main}) comes from
%generalization, not from pseudo-label quality alone.

%
%We additionally evaluated DropLoss~\cite{cups}, which suppresses gradient from
%thing predictions unmatched by any pseudo-label instance. On RPTU it did not
%improve results: at convergence, DropLoss slightly reduced thing quality
%(PQ$_{\text{Th}}$ $29.21 \rightarrow 28.76$) while leaving stuff unchanged, and
%we therefore report the standard-loss model as our main result.

%\paragraph{Evaluation-set limitation.}
%The $15$ GT images serve as both cluster-mapping calibration and ablation
%evaluation, so these pseudo-label numbers should be read as a controlled
%component comparison rather than a held-out generalization estimate. The
%downstream Mask2Former result in Sec.~\ref{sec:rptu_main} is the more
%meaningful measure of pipeline quality, as the model never sees GT during
%training.

\subsection{Edge Case Evaluation}
\label{sec:finnwoodlands}

Additionally, we evaluate the performance of ProGuT's panoptic track on Finnwoodlands which showcases a much difficult domain than Our forest. Finnwoodlands dataset has winter forest with snow-occluded trees. We evaluate on $50$ validation images, remapped to four coarse classes (\ref{sec:datasets}), using the same pseudo-labels pipeline as Our forest. 

%FinnWoodlands is the only publicly available forest panoptic benchmark, and we
%include it as a stress test of a qualitatively harder regime than RPTU: a
%Finnish winter forest in which snow occludes the lower $40$--$60\%$ of most
%trunks. We evaluate on the $50$ public validation images, remapped to four
%coarse classes (Sec.~\ref{sec:datasets}). The pseudo-label trunk masks are
%generated with the same pipeline as RPTU; the FinnWoodlands evaluation revealed
%a labeling bug in our earlier protocol (the GT \texttt{segments\_info} lacks an
%\texttt{isthing} field, silently discarding all trunk instances), which we fixed
%by keying thing categories explicitly; the numbers below are post-fix.
\begin{figure}[!htp]
	\centering
	\includegraphics[width=\linewidth]{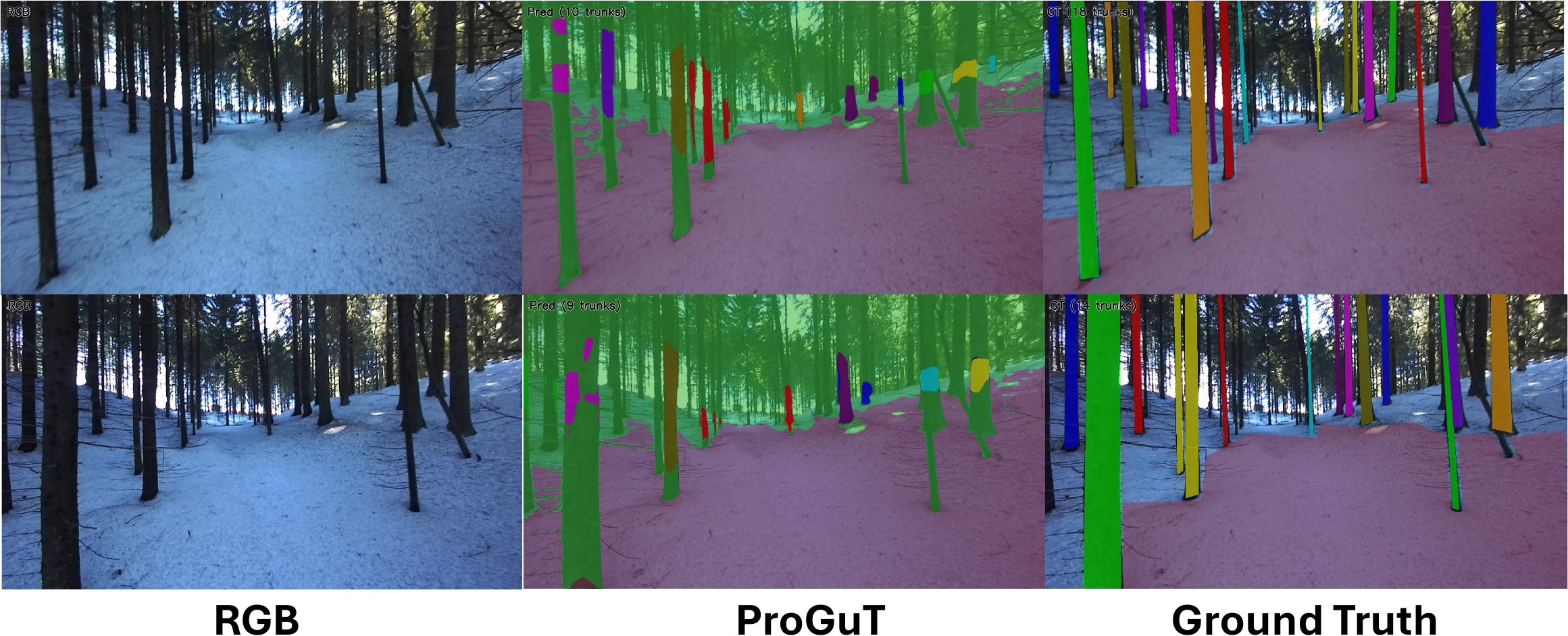}
	\caption{Qualitative result of ProGuT on FinnWoodlands \cite{lagos2023finnwoodlands} dataset}
	\label{fig:panoptic_track_finnwoodlands}
\end{figure}
\begin{table}[t]
	\centering
	\caption{Panoptic segmentation on FinnWoodlands val set. Pseudo-label modes are evaluated directly against 4-class coarse GT; ProGuT + Mask2Former denotes the downstream model.}
	\label{tab:finnwoodlands}
	\scriptsize
	\setlength{\tabcolsep}{2.5pt}
	\renewcommand{\arraystretch}{1.08}
	\begin{tabularx}{\columnwidth}{@{}>{\raggedright\arraybackslash}Xcccc@{}}
		\toprule
		\textbf{Method} & \textbf{PQ} & \textbf{PQ$_{\mathrm{Th}}$} &\textbf{ PQ$_{\mathrm{St}}$} & \textbf{\shortstack{Ground\\PQ}} \\
		\midrule
		semantic-only & 9.10 & 0.00 & 52.29 & 79.71 \\
		coherence-only & 6.09 & 0.09 & 52.29 & 79.71 \\
		unet-only & 7.00 & 0.11 & 52.29 & 79.71 \\
		dual-pass & 5.42 & 0.15 & 52.29 & 79.71 \\
		\midrule
		\textbf{ProGuT + Mask2Former} & \textbf{8.17} & \textbf{1.33} & \textbf{53.14} & \textbf{79.71} \\
		\bottomrule
	\end{tabularx}
\end{table}

As seen from Table.~\ref{tab:finnwoodlands}, at IoU$ = 0.50$, the model matches only $12$ of the $712$ GT trunks ($\text{tp}=12$, $\text{fp}=262$, $\text{fn}=700$); at a relaxed threshold (IoU $=0.08$), it matches $52\%$ (RQ $=52.33\%$). We argue that trunks detection are spatially correct but vertically truncated (as seen from Figure.~\ref{fig:panoptic_track_finnwoodlands}). The stuff component however, is unaffected; ground is present in every image, recovered at Ground PQ $=79.71$, RQ $=100\%$.

%% file: sec/conclusion.tex
\section{Conclusion}
\label{sec:conclusion}

We opened by asking what a robot must answer before navigating a forest: \emph{\textit{\textbf{where can it drive ?}}} and \emph{\textit{\textbf{what does it see around itself ?}}}. With ProGuT we answer both without heavy supervision. For segregating regions (\textbf{stuff}) (for e.g. to find a driveable path), the pseudo-labels alone train a downstream models to \textbf{competitive accuracy $65.89$ mIoU} on Freiburg Forest, exceeding STEGO ($57.57$) and PiCIE ($45.25$) thereby making ProGuT as a data engine in order to substitute for manual annotations. In a more challenging task, to separate instances (\textbf{things }which the robot sees around itself), we show that simple geometrical falsification principle can be more effective than the current saliency and motion-based object discovery methods (U2Seg reaching PQ=$3.75$ with zero thing quality). \\
Future directions include using vision-language alignment to automatically ground clusters to classes, extending the geometric falsification principle to a broad set of forest/off-road classes.     

%% file: main.bib
@String(ICCV= {Int. Conf. Comput. Vis.})

@String(ICCV  = {ICCV})

@inproceedings{lagos2023finnwoodlands,
  title={Finnwoodlands dataset},
  author={Lagos, Juan and Lempi{\"o}, Urho and Rahtu, Esa},
  booktitle={Scandinavian Conference on Image Analysis},
  pages={95--110},
  year={2023},
  organization={Springer}
}

@article{hamilton2022unsupervised,
  title={Unsupervised semantic segmentation by distilling feature correspondences},
  author={Hamilton, Mark and Zhang, Zhoutong and Hariharan, Bharath and Snavely, Noah and Freeman, William T},
  journal={arXiv preprint arXiv:2203.08414},
  year={2022}
}

@article{zhang2022dino,
  title={Dino: Detr with improved denoising anchor boxes for end-to-end object detection},
  author={Zhang, Hao and Li, Feng and Liu, Shilong and Zhang, Lei and Su, Hang and Zhu, Jun and Ni, Lionel M and Shum, Heung-Yeung},
  journal={arXiv preprint arXiv:2203.03605},
  year={2022}
}

@article{zadaianchuk2022unsupervised,
  title={Unsupervised semantic segmentation with self-supervised object-centric representations},
  author={Zadaianchuk, Andrii and Kleindessner, Matthaeus and Zhu, Yi and Locatello, Francesco and Brox, Thomas},
  journal={arXiv preprint arXiv:2207.05027},
  year={2022}
}

@inproceedings{sick2024unsupervised,
  title={Unsupervised semantic segmentation through depth-guided feature correlation and sampling},
  author={Sick, Leon and Engel, Dominik and Hermosilla, Pedro and Ropinski, Timo},
  booktitle={Proceedings of the IEEE/CVF Conference on Computer Vision and Pattern Recognition},
  pages={3637--3646},
  year={2024}
}

@ARTICLE{868688,
  author={Jianbo Shi and Malik, J.},
  journal={IEEE Transactions on Pattern Analysis and Machine Intelligence}, 
  title={Normalized cuts and image segmentation}, 
  year={2000},
  volume={22},
  number={8},
  pages={888-905},
  doi={10.1109/34.868688}}

@inproceedings{valada2017icra,
  author = {Valada, Abhinav and Vertens, Johan and Dhall, Ankit and Burgard, Wolfram},
  title = {AdapNet: Adaptive Semantic Segmentation in Adverse Environmental Conditions},
  booktitle = {Proceedings of the IEEE International Conference on Robotics and Automation (ICRA)},
  pages={4644--4651},
  year = {2017},
  organization={IEEE}
}

@inproceedings{wang2023cut,
  title={Cut and learn for unsupervised object detection and instance segmentation},
  author={Wang, Xudong and Girdhar, Rohit and Yu, Stella X and Misra, Ishan},
  booktitle={Proceedings of the IEEE/CVF conference on computer vision and pattern recognition},
  pages={3124--3134},
  year={2023}
}

@article{edlinger2023terrain,
  title={Terrain segmentation for commercial vehicles and working machines},
  author={Edlinger, Raimund and Mitterhuber, Ulrich and N{\"u}chter, Andreas},
  journal={Electronic Imaging},
  volume={35},
  pages={1--7},
  year={2023},
  publisher={Society for Imaging Science and Technology}
}

@misc{oliveira2020singlestageencoderdecodernetworks,
      title={Beyond Single Stage Encoder-Decoder Networks: Deep Decoders for Semantic Image Segmentation}, 
      author={Gabriel L. Oliveira and Senthil Yogamani and Wolfram Burgard and Thomas Brox},
      year={2020},
      eprint={2007.09746},
      archivePrefix={arXiv},
      primaryClass={cs.CV},
      url={https://arxiv.org/abs/2007.09746}, 
}

@inproceedings{arica2024cuvler,
  title={CuVLER: Enhanced unsupervised object discoveries through exhaustive self-supervised transformers},
  author={Arica, Shahaf and Rubin, Or and Gershov, Sapir and Laufer, Shlomi},
  booktitle={Proceedings of the IEEE/CVF Conference on Computer Vision and Pattern Recognition},
  pages={23105--23114},
  year={2024}
}

@inproceedings{wigness2019rugd,
  title={A rugd dataset for autonomous navigation and visual perception in unstructured outdoor environments},
  author={Wigness, Maggie and Eum, Sungmin and Rogers, John G and Han, David and Kwon, Heesung},
  booktitle={2019 IEEE/RSJ International Conference on Intelligent Robots and Systems (IROS)},
  pages={5000--5007},
  year={2019},
  organization={IEEE}
}

@inproceedings{jiang2021rellis,
  title={Rellis-3d dataset: Data, benchmarks and analysis},
  author={Jiang, Peng and Osteen, Philip and Wigness, Maggie and Saripalli, Srikanth},
  booktitle={2021 IEEE international conference on robotics and automation (ICRA)},
  pages={1110--1116},
  year={2021},
  organization={IEEE}
}

@inproceedings{mortimer2024goose,
  title={The goose dataset for perception in unstructured environments},
  author={Mortimer, Peter and Hagmanns, Raphael and Granero, Miguel and Luettel, Thorsten and Petereit, Janko and Wuensche, Hans-Joachim},
  booktitle={2024 IEEE International Conference on Robotics and Automation (ICRA)},
  pages={14838--14844},
  year={2024},
  organization={IEEE}
}

@article{jiang2025go,
  title={Go: The great outdoors multimodal dataset},
  author={Jiang, Peng and Viswanath, Kasi and Nagariya, Akhil and Chustz, George and Wigness, Maggie and Osteen, Philip and Overbye, Timothy and Ellis, Christian and Quang, Long and Saripalli, Srikanth},
  journal={arXiv preprint arXiv:2501.19274},
  year={2025}
}

@InProceedings{valada16iser,
author = {Abhinav Valada and Gabriel Oliveira and Thomas Brox and Wolfram Burgard},
title = {Deep Multispectral Semantic Scene Understanding of Forested Environments using Multimodal Fusion},
booktitle = {International Symposium on Experimental Robotics (ISER)},
year = {2016},
}

@article{vidanapathirana2025wildscenes,
  title={Wildscenes: A benchmark for 2d and 3d semantic segmentation in large-scale natural environments},
  author={Vidanapathirana, Kavisha and Knights, Joshua and Hausler, Stephen and Cox, Mark and Ramezani, Milad and Jooste, Jason and Griffiths, Ethan and Mohamed, Shaheer and Sridharan, Sridha and Fookes, Clinton and others},
  journal={The International Journal of Robotics Research},
  volume={44},
  number={4},
  pages={532--549},
  year={2025},
  publisher={Sage Publications Sage UK: London, England}
}

@article{grondin2022tree,
    author = {Grondin, Vincent and Fortin, Jean-Michel and Pomerleau, François and Giguère, Philippe},
    title = {Tree detection and diameter estimation based on deep learning},
    journal = {Forestry: An International Journal of Forest Research},
    year = {2022},
    month = {10},
}

@article{da2021visible,
  title={Visible and thermal image-based trunk detection with deep learning for forestry mobile robotics},
  author={Da Silva, Daniel Queir{\'o}s and Dos Santos, Filipe Neves and Sousa, Armando Jorge and Filipe, V{\'\i}tor},
  journal={Journal of imaging},
  volume={7},
  number={9},
  pages={176},
  year={2021},
  publisher={MDPI}
}

@article{henrich2024treelearn,
  title={TreeLearn: A deep learning method for segmenting individual trees from ground-based LiDAR forest point clouds},
  author={Henrich, Jonathan and van Delden, Jan and Seidel, Dominik and Kneib, Thomas and Ecker, Alexander S},
  journal={Ecological Informatics},
  volume={84},
  pages={102888},
  year={2024},
  publisher={Elsevier}
}

@inproceedings{ji2019invariant,
  title={Invariant information clustering for unsupervised image classification and segmentation},
  author={Ji, Xu and Henriques, Joao F and Vedaldi, Andrea},
  booktitle={Proceedings of the IEEE/CVF international conference on computer vision},
  pages={9865--9874},
  year={2019}
}

@inproceedings{cho2021picie,
  title={Picie: Unsupervised semantic segmentation using invariance and equivariance in clustering},
  author={Cho, Jang Hyun and Mall, Utkarsh and Bala, Kavita and Hariharan, Bharath},
  booktitle={Proceedings of the IEEE/CVF conference on computer vision and pattern recognition},
  pages={16794--16804},
  year={2021}
}

@article{li2022language,
  title={Language-driven semantic segmentation},
  author={Li, Boyi and Weinberger, Kilian Q and Belongie, Serge and Koltun, Vladlen and Ranftl, Ren{\'e}},
  journal={arXiv preprint arXiv:2201.03546},
  year={2022}
}

@inproceedings{dong2023maskclip,
  title={Maskclip: Masked self-distillation advances contrastive language-image pretraining},
  author={Dong, Xiaoyi and Bao, Jianmin and Zheng, Yinglin and Zhang, Ting and Chen, Dongdong and Yang, Hao and Zeng, Ming and Zhang, Weiming and Yuan, Lu and Chen, Dong and others},
  booktitle={Proceedings of the IEEE/CVF conference on computer vision and pattern recognition},
  pages={10995--11005},
  year={2023}
}

@inproceedings{ghiasi2022scaling,
  title={Scaling open-vocabulary image segmentation with image-level labels},
  author={Ghiasi, Golnaz and Gu, Xiuye and Cui, Yin and Lin, Tsung-Yi},
  booktitle={European conference on computer vision},
  pages={540--557},
  year={2022},
  organization={Springer}
}

@inproceedings{cheng2021maskformer,
  title={Per-Pixel Classification is Not All You Need for Semantic Segmentation},
  author={Bowen Cheng and Alexander G. Schwing and Alexander Kirillov},
  journal={NeurIPS},
  year={2021}
}

@inproceedings{cheng2022masked,
  title={Masked-attention mask transformer for universal image segmentation},
  author={Cheng, Bowen and Misra, Ishan and Schwing, Alexander G and Kirillov, Alexander and Girdhar, Rohit},
  booktitle={Proceedings of the IEEE/CVF conference on computer vision and pattern recognition},
  pages={1290--1299},
  year={2022}
}

@misc{wang2021detectingmappingtreesunstructured,
      title={Detecting and Mapping Trees in Unstructured Environments with a Stereo Camera and Pseudo-Lidar}, 
      author={Brian H. Wang and Carlos Diaz-Ruiz and Jacopo Banfi and Mark Campbell},
      year={2021},
      eprint={2103.15967},
      archivePrefix={arXiv},
      primaryClass={cs.RO},
      url={https://arxiv.org/abs/2103.15967}, 
}

@inproceedings{niu2024unsupervised,
  title={Unsupervised universal image segmentation},
  author={Niu, Dantong and Wang, Xudong and Han, Xinyang and Lian, Long and Herzig, Roei and Darrell, Trevor},
  booktitle={Proceedings of the IEEE/CVF conference on computer vision and pattern recognition},
  pages={22744--22754},
  year={2024}
}

@inproceedings{hahn2025scene,
  title={Scene-centric unsupervised panoptic segmentation},
  author={Hahn, Oliver and Reich, Christoph and Araslanov, Nikita and Cremers, Daniel and Rupprecht, Christian and Roth, Stefan},
  booktitle={Proceedings of the Computer Vision and Pattern Recognition Conference},
  pages={24485--24495},
  year={2025}
}

@inproceedings{kirillov2023segment,
  title={Segment anything},
  author={Kirillov, Alexander and Mintun, Eric and Ravi, Nikhila and Mao, Hanzi and Rolland, Chloe and Gustafson, Laura and Xiao, Tete and Whitehead, Spencer and Berg, Alexander C and Lo, Wan-Yen and others},
  booktitle={Proceedings of the IEEE/CVF international conference on computer vision},
  pages={4015--4026},
  year={2023}
}

@inproceedings{ravi2025sam,
  title={Sam 2: Segment anything in images and videos},
  author={Ravi, Nikhila and Gabeur, Valentin and Hu, Yuan-Ting and Hu, Ronghang and Ryali, Chaitanya and Ma, Tengyu and Khedr, Haitham and R{\"a}dle, Roman and Rolland, Chloe and Gustafson, Laura and others},
  booktitle={International Conference on Learning Representations},
  volume={2025},
  pages={28085--28128},
  year={2025}
}

@inproceedings{forstner1987fast,
  author    = {F{\"{o}}rstner, Wolfgang and G{\"{u}}lch, Eberhard},
  title     = {A Fast Operator for Detection and Precise Location of Distinct Points},
  booktitle = {Proceedings of the Intercommission Conference on Fast Processing of Photogrammetric Data},
  pages     = {281--305},
  year      = {1987},
  address   = {Interlaken, Switzerland}
}

@inproceedings{bigun1987optimal,
  author    = {Big{\"{u}}n, Josef and Granlund, G{\"{o}}sta H.},
  title     = {Optimal Orientation Detection of Linear Symmetry},
  booktitle = {Proceedings of the First IEEE International Conference on Computer Vision (ICCV)},
  pages     = {433--438},
  year      = {1987},
  address   = {London, UK}
}

@inproceedings{harris1988combined,
  author    = {Harris, Chris and Stephens, Mike},
  title     = {A Combined Corner and Edge Detector},
  booktitle = {Proceedings of the 4th Alvey Vision Conference},
  pages     = {147--151},
  year      = {1988},
  volume    = {15},
  address   = {Manchester, UK}
}

@article{weickert1999coherence,
  author    = {Weickert, Joachim},
  title     = {Coherence-Enhancing Diffusion Filtering},
  journal   = {International Journal of Computer Vision},
  volume    = {31},
  number    = {2-3},
  pages     = {111--127},
  year      = {1999},
  publisher = {Springer}
}

@article{weickert1999coherence_color,
  author    = {Weickert, Joachim},
  title     = {Coherence-enhancing diffusion of colour images},
  journal   = {Image and Vision Computing},
  volume    = {17},
  number    = {3-4},
  pages     = {199--210},
  year      = {1999},
  publisher = {Elsevier}
}

@article{liu2025data,
  title={Data or Language Supervision: What Makes CLIP Better than DINO?},
  author={Liu, Yiming and Zhang, Yuhui and Ghosh, Dhruba and Schmidt, Ludwig and Yeung-Levy, Serena},
  journal={arXiv preprint arXiv:2510.11835},
  year={2025}
}
